\documentclass[runningheads]{llncs}
\usepackage[T1]{fontenc}
\usepackage{graphicx}
\usepackage{booktabs}
\usepackage{hyperref}
\usepackage[misc]{ifsym}
\newcommand{\corr}{(\Letter)}
\usepackage{mwe}
\usepackage{hyperref} 

\begin{document}

\title{Volvo Discovery Challenge at ECML-PKDD 2024}


\author{Mahmoud Rahat\inst{1} \corr \and 
Peyman Sheikholharam Mashhadi \inst{1} \and 
S\l{}awomir Nowaczyk \inst{1}  \and
Shamik Choudhury \inst{2} \and
Leo Petrin \inst{2},
Thorsteinn Rognvaldsson \inst{1}, Andreas Voskou \inst{3}, Carlo Metta \inst{4} \and Claudio Savelli \inst{5} }

\authorrunning{Rahat et al., 2024}

\institute{Center for Applied Intelligent Systems Research, Halmstad University, Sweden \email{\{mahmoud.rahat,peyman.mashhadi,slawomir.nowaczyk,thorsteinn.rognvaldsson\}@hh.se}
\and
Volvo Group Trucks Technology, Sweden \email{shamik.choudhury@consultant.volvo.com, leo.petrin@volvo.com}
\and  Boltzmann Research, Cyprus, \email{andreas@boltzmann-research.com}
\and ISTI-CNR, Italy, \email{carlo.metta@isti.cnr.it}
\and Politecnico di Torino, Italy, \email{claudio.savelli@studenti.polito.it}}

\maketitle              

\begin{abstract}
This paper presents an overview of the Volvo Discovery Challenge, held during the ECML-PKDD 2024 conference. The challenge's goal was to predict the failure risk of an anonymized component in Volvo trucks using a newly published dataset. The test data included observations from two generations (gen1 and gen2) of the component, while the training data was provided only for gen1. The challenge attracted 52 data scientists from around the world who submitted a total of 791 entries. We provide a brief description of the problem definition, challenge setup, and statistics about the submissions. In the section on winning methodologies, the first, second, and third-place winners of the competition briefly describe their proposed methods and provide GitHub links to their implemented code. The shared code can be interesting as an advanced methodology for researchers in the predictive maintenance domain. The competition was hosted on the Codabench platform\footnote{\href{https://www.codabench.org/competitions/3022/?secret_key=c5bb4004-b280-456e-84f6-3bb42737e8dc}{Link} to the challenge page.}.

\end{abstract}

\section{Introduction}
The Volvo Discovery Challenge was held as part of the European Conference on Machine Learning and Data Mining \href{https://ecmlpkdd.org/2024/}{(ECML PKDD 2024)}. In collaboration with \href{https://www.volvotrucks.com/en-en/}{Volvo Group Truck Technologies}, \href{https://www.hh.se/}{Halmstad University} challenged participants to predict the risk of failure for an undisclosed component of Volvo trucks. This challenge invited participants to work with an exclusive real-world dataset containing measurements from a fleet of more than 10,000 Volvo heavy-duty trucks. The task was to develop a machine learning model to predict risk levels (i.e. \textit{Low}, Medium and \textit{High}) for a component of the trucks in the test set. The core task of this competition falls within the broader scope of predictive maintenance (PdM).the

PdM aims to use data to forecast equipment failures, reduce downtime, and make maintenance more intelligent. The solution to this competition could enhance the  PdM strategies for vehicles, improving reliability and efficiency and reducing the environmental impact and $CO_2$ emission. For some operational reasons, we decided to formulate the challenge as a classification task, but a review of the literature reveals that this problem can be formulated in at least three distinct ways: classification \cite{altarabichi2020stacking}, regression\cite{rahat2023bridging}, and survival analysis \cite{rahat2024survloss}. Each one of the mentioned directions has its own benefits and drawbacks. 

The challenge participants investigated various techniques from these three directions and devised innovative solutions to improve their prediction scores. As an example, when we look at the history of the submissions, we can see that the initial submissions of the first-place winner revolved around a score of 0.53. Submitting 118 predictions during 7 weeks improved this number to 0.89. You can read a short description of the top three methodologies in section \ref{sec:methodology}. The rest of the paper describes more details about the challenge, submissions, and dataset description. \href{https://halmstaduniversity.box.com/s/x2e8gfcb37an77wwc566ror3r3sg2yg1}{Here}, you can find the challenge description file that was shared with the participants.

\section{Challenge Setup}

The challenge was hosted on the Codabench platform (\href{https://www.codabench.org/competitions/3022/?secret_key=c5bb4004-b280-456e-84f6-3bb42737e8dc}{Link}) and had two phases:
\begin{enumerate}
    \item \textbf{Development Phase} (May 15 to June 15, 2024): In the development phase, the participants could submit 5 predictions per day. The submitted prediction was evaluated against 20 percent of the ground truth data. The best-performing submission from each contestant was automatically published on the leaderboard, allowing everyone to compare their results with others.
    \item \textbf{Final Phase} (June 16 to June 30, 2024): The final phase started after the development phase. In this phase, the contestants were allowed to submit only 3 predictions in total. The submitted prediction was evaluated against the whole ground truth data. The best-performing submission among those three submissions in the final phase was displayed on the Leaderboard. The ranking in the Leaderboard of the final phase determined the final ranking.
\end{enumerate}

\subsection{Prizes}

The top three competitors received a time slot to present their work at the conference. The first place received a free registration for the ECML-PKDD 2024. In addition, there was prize money for the top three contenders: the team in the first place received 500€, the second place 300€, and the third place 200€.

\subsection{Submission Process}

 Each submission was a single ZIP file called \textit{prediction.zip}. Inside the file, there was a CSV file containing a header called \textit{pred} followed by one prediction (Low, Medium, or High) per row for a total of 33590 rows. The index of each row in the prediction file was aligned with the index of the \textit{public\_X\_test.csv} file. The participants were asked to submit the \textit{prediction.zip} file in the Codabench portal. Then, a Python code developed by the organizers (\href{https://github.com/mahmoudrahat/VolvoChallengeECML-PKDD2024/blob/main/scoring.py}{link}) automatically evaluates the prediction and returns the scores.

To make the submission process easy to follow, the participants were provided with a start kit (\href{https://github.com/mahmoudrahat/VolvoChallengeECML-PKDD2024/blob/main/startkit.py}{link}). This file served as a starting point by first reading the training data, then training a Decision Tree classifier as a baseline, performing predictions on the test set, and finally creating the \textit{prediction.zip} file. The produced file can be submitted directly to the Codabench portal. An example of a submission file is shared \href{https://github.com/mahmoudrahat/VolvoChallengeECML-PKDD2024/blob/main/SampleSubmission.zip}{here}.

\section{Submissions Statistics}

\begin{figure}[h!]
\centering
\includegraphics[width=1\linewidth]{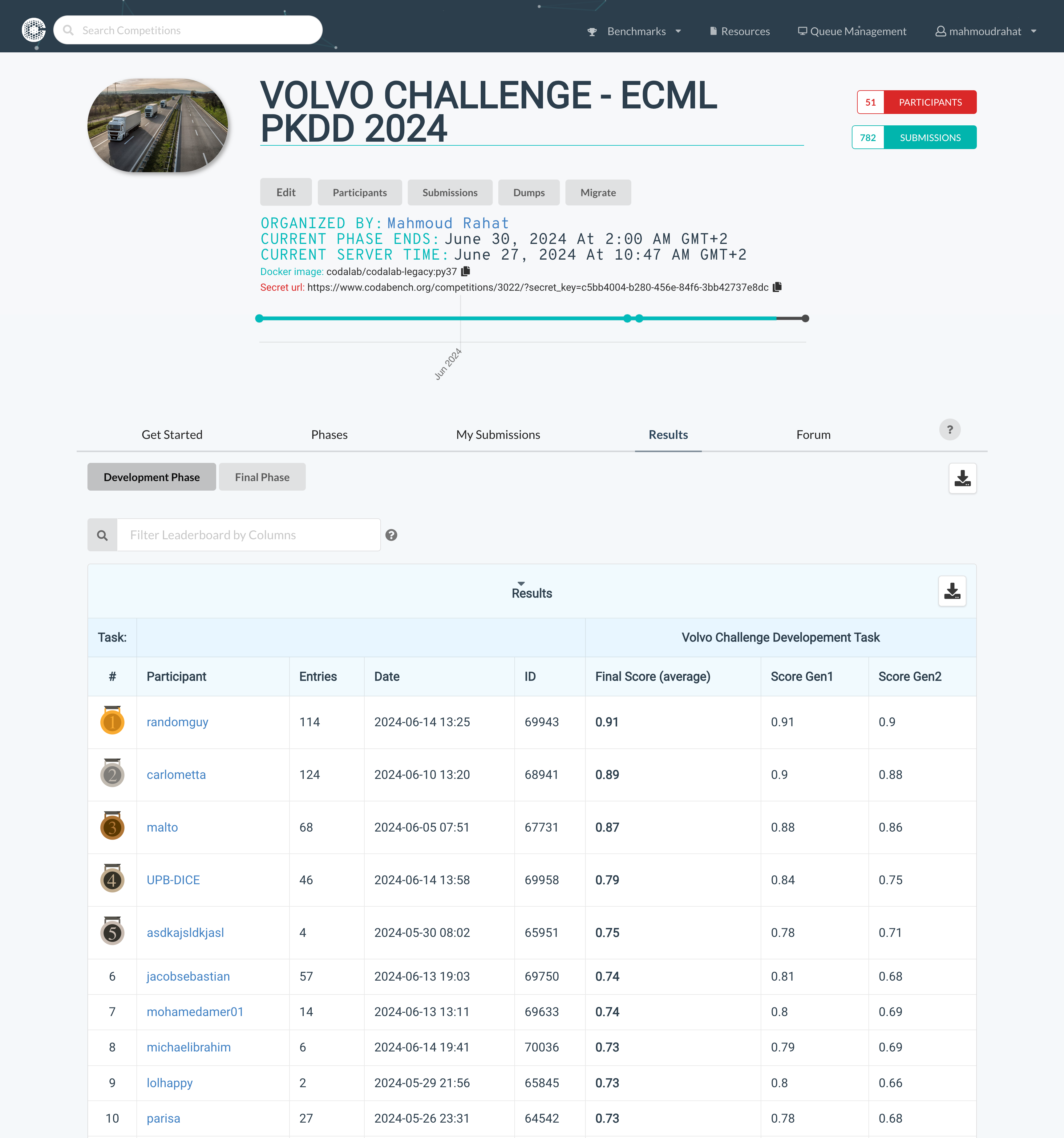}
\caption{\label{fig:developmentPhase}Development phase leaderboard for the top ten places.}
\end{figure}

\begin{figure}[h!]
\centering
\includegraphics[width=1\linewidth]{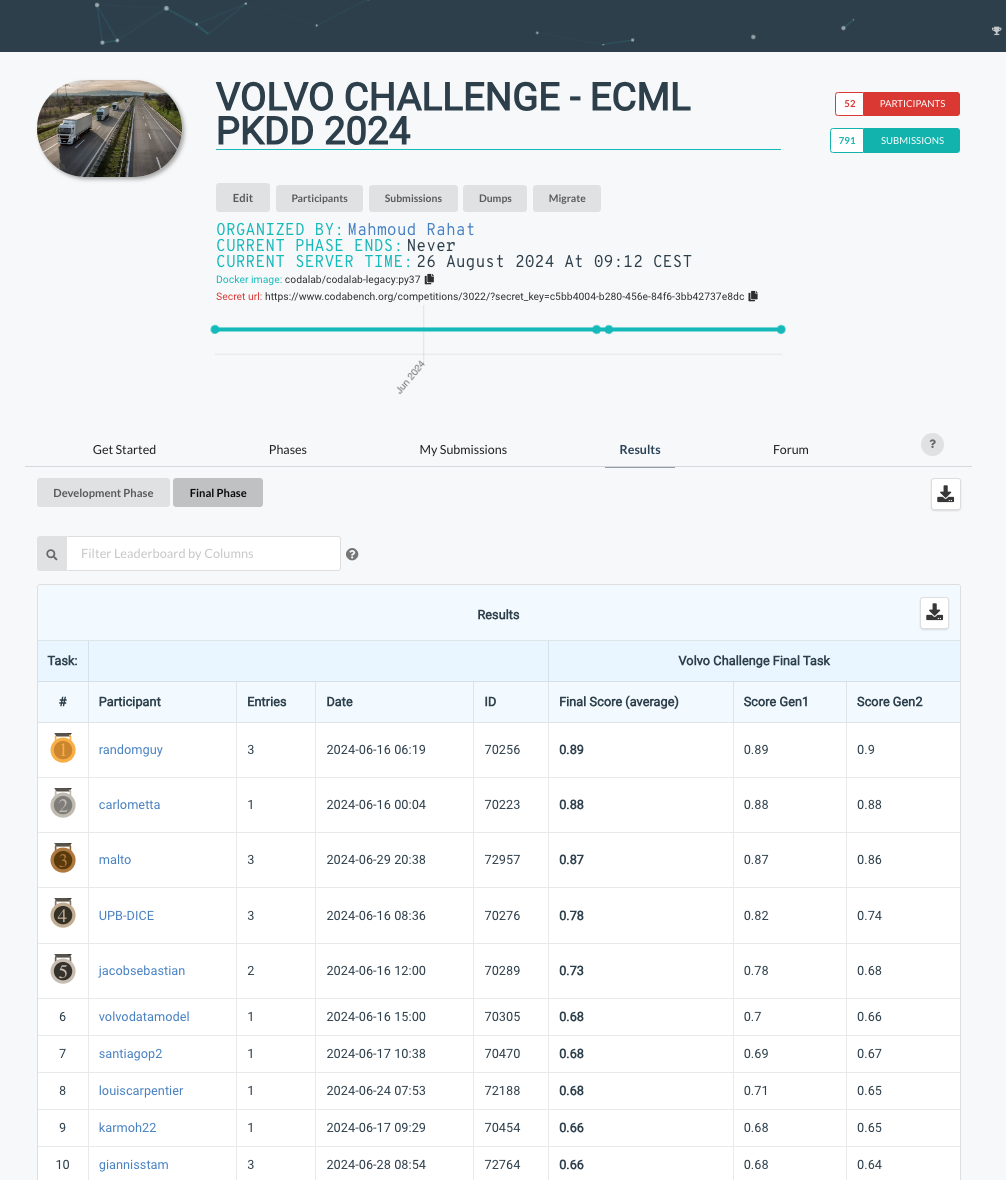}
\caption{\label{fig:finalPhase}Final phase leaderboard for the top ten places.}
\end{figure}

The competition received a total of 791 submissions from 52 participants, with 28 submissions in the final phase and 763 in the development phase. Figures \ref{fig:developmentPhase} and \ref{fig:finalPhase} show leaderboards in the Codabench portal for the top 10 positions of the development and final phases, respectively. As shown in the figures, the top three positions remained unchanged between the two phases. The performance of the models decreased by only about 2 percent when moving from the development to the final phase. This indicates the models had a high level of generalization ability. Another interesting observation is the low difference (less than 1 percent) between the performance of the models in \textit{gen1} compared to \textit{gen2}.

\begin{figure}[h!]
\centering
\includegraphics[width=0.8\linewidth]{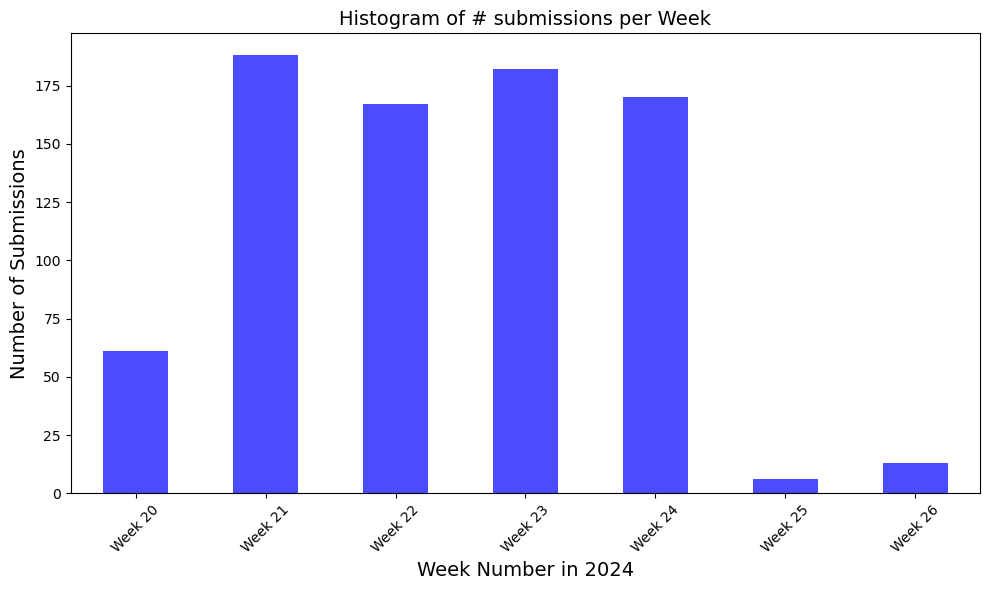}
\caption{\label{fig:histogram}A histogram showing number of submissions per week.}
\end{figure}

\begin{figure}[h!]
\centering
\includegraphics[width=0.8\linewidth]{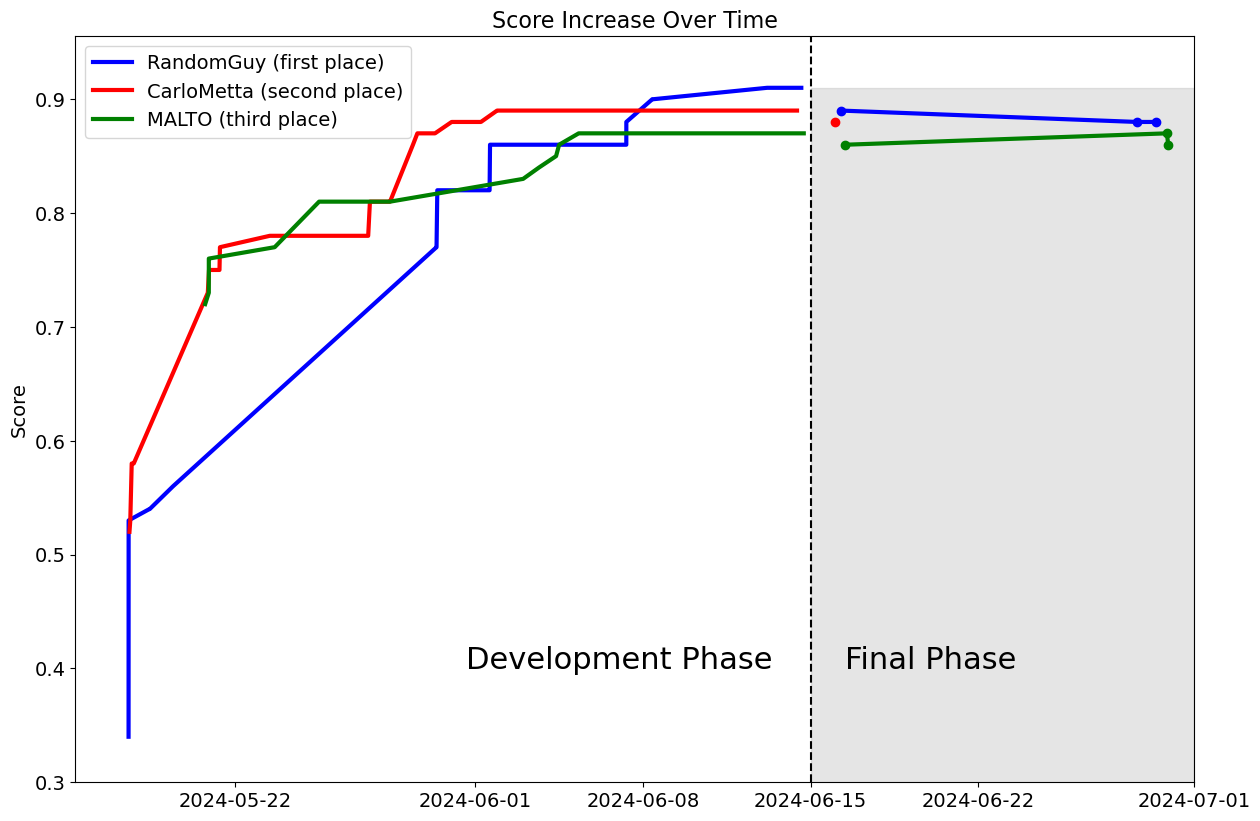}
\caption{\label{fig:scoresOverTime}Score improvement over time for the top three participants.}
\end{figure}

The competition ran for 6 weeks between week 20 to week 26. Figure \ref{fig:histogram}  shows  the total number of submissions per week. The last two weeks show the final phase, and fewer submissions were received. Ultimately, Figure \ref{fig:scoresOverTime} visualizes the score improvement over time for the top three places. To produce this figure, submissions are displayed only if they received a higher grade than the best score seen so far.

\section{Description of the Dataset}

The dataset contains three files: \texttt{train\_gen1.csv}, \texttt{public\_x\_test.csv}, and \texttt{variants.csv}. The \texttt{train\_gen1.csv} file contains 157,437 readouts from an undisclosed component across 7,280 Volvo heavy-duty trucks, each uniquely identified by an anonymous chassis ID labeled \texttt{ChassisId\_encoded}. This file includes 308 columns. The first four columns are \texttt{Timesteps}, \texttt{ChassisId\_encoded}, \texttt{gen}, and  \texttt{risk\_level}, followed by 304 feature columns. Note that the \texttt{train\_gen1.csv} file only contains data from the first generation of the component.  These readouts are recorded at consecutive timesteps, starting from timestep 1 and extending until either component failure or the end of data collection. While some components experienced failure during the study period (unhealthy components), the majority did not fail (healthy components). The time intervals between the consecutive readouts are undisclosed but can be considered consistent and equally spaced.

\begin{figure}[h!]
\centering
\includegraphics[width=0.5\linewidth]{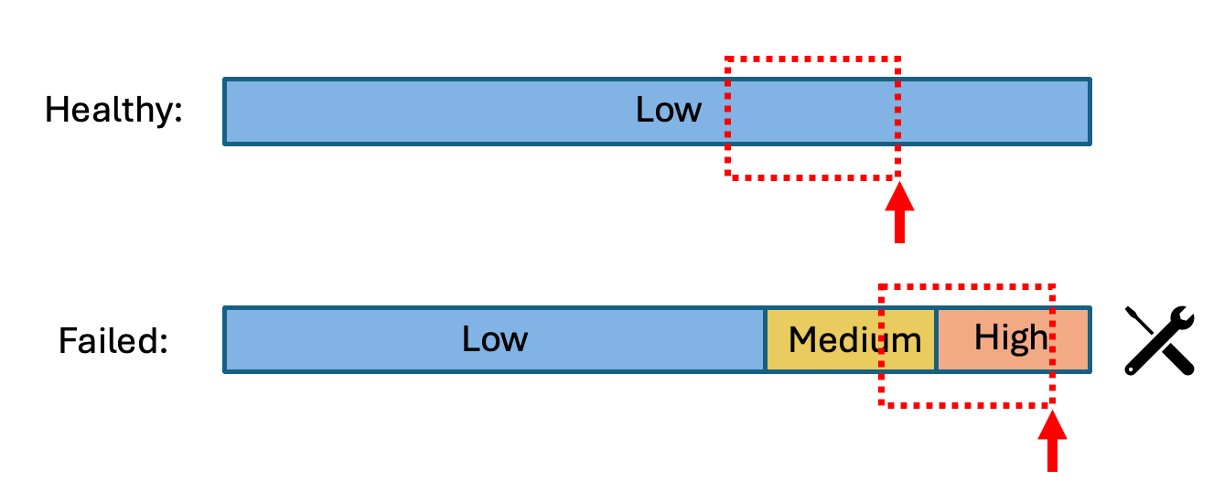}
\caption{\label{fig:label}Sequence Extraction from Healthy and Failed components. Note that the length of all sequences is 10 time steps.}
\end{figure}

The training set includes the target variable \texttt{risk\_level}, categorizing the risk for trucks into three levels: \textit{Low}, \textit{Medium}, and \textit{High}. Each readout is assigned one of these labels based on its proximity to the component failure time. As can be seen in Figure \ref{fig:label}, The \textit{High} label is assigned to readouts within 9 timesteps before a failure, the \textit{Medium} label to those between 9 and 18 timesteps, and finally, the \textit{Low} label to readouts that are at least 18 timesteps away from failure. Additionally, it is important to mention that the training set was limited to observations from a single generation, called \textit{gen1}, of the component. However, the competition's task extended beyond this scope, and the model was expected to provide predictions for both generations of components (\textit{gen1} and \textit{gen2}) in the test dataset. The goal was to evaluate how well a model can perform on the test data from gen1 and also how well it can generalize to a new, unseen generation, i.e., gen2.


\texttt{public\_X\_test.csv} includes 33590 rows and 307 columns where the first three columns are \texttt{Timesteps}, \texttt{ChassisId\_encoded}, \texttt{gen}, and the remaining 304 columns are sensor readings, mirroring the features that exist in the training set. Unlike the training set, the test data spans two generations of the case study's component (as indicated in the \texttt{gen} column), adding more complexity to the prediction task. The prediction file should include one prediction per row of \texttt{public\_X\_test} file (i.e., 33590 rows).
Moreover, in the test set, the sequence length from each chassis ID is fixed at 10 Timesteps. For the healthy components, a sequence with a length of 10 is randomly selected from each \texttt{ChassisId\_encoded}. In failure cases, a random timestep from the \texttt{High risk} period is selected as the last readout, with the preceding 10 Timesteps initiating the sequence for that particular test chassis ID. The sequence selection process is illustrated in Figure \ref{fig:label}.

Finally, \texttt{variants.csv} contains specifications for all trucks included in the \textit{train\_gen1.csv} and \textit{public\_X\_test.csv} files. The file has 10,639 rows and 13 columns. The first column gives \texttt{ChassisId\_encoded}, followed by 12 columns containing the encoded specifications of the trucks. The specifications include information such as the engine type, cabin type, number of wheels, number of axles, etc.

\section{Evaluation}

The macro-average F1-score was the main evaluation metric in this competition. To calculate the macro-average F1-score for a classifier with three classes (i.e., Low, Medium, High), we first compute the F1-score for each class individually and then take the average of these F1-scores. The macro-average F1-score treats all classes equally, regardless of class imbalance.

The macro-average F1-score is calculated and reported for predictions of gen1 and gen2 separately. The final score determining the winner was the average of the macro F1-score for gen1 and gen2. The startkit.py file contains an implementation of the evaluation metric used in the competition and evaluates the predictions of the baseline model against a mocked-up ground truth.

\section{Winning Methodologies} \label{sec:methodology}

In this section, we provide a brief overview of the methods employed by the first, second, and third place winners in the competition, respectively, presented in the below subsections.

\subsection{Team RandomGuy}
The proposed solution\footnote{https://github.com/avoskou/Volvo-Challenge} employs a tabular data classification approach, analyzing each row of input data individually with minimal involvement of temporal features. It follows a two-step process: first, identifying whether a row belongs to a healthy or non-healthy sequence, and second, reclassifying the non-healthy elements as "Medium" or "High" risk.

In the first step, each data row is classified to determine whether it belongs to a healthy (low-risk) sequence. This classification is achieved using the recent STab model \cite{voskou2024transformers}, which is based on a stochastic transformer architecture specifically designed for tabular data. The model is trained using all primary input features, except for the "Variants" features, and employs binary labels to indicate whether a row is healthy or not. Healthy rows are defined as those from sequences that remain low-risk throughout, while non-healthy rows are selected only if they are classified as medium or high risk.

An ensemble of five STab models is trained on different balanced bootstrap samples of the training data, where rare labels are randomly upsampled. During inference, each of the five models is run 20 times due to  the stochastic nature of the STab model, resulting in  100 total predictions per row. The average of these predictions is used to assign a probability score to each row, and specific thresholds for the minimum, maximum, and mean of the scores across all rows in a sequence are then set to make a final classification. For sequences classified as healthy, all predictions are set to "Low". For those classified as non-healthy, the process advances to the next step.

Although the first model demonstrates high accuracy, the second subtask initially shows weak predictive performance. To address this, the solution adopts a more manual approach, starting by classifying the last seven elements of a sequence as "High" and the first three as "Medium", which establishes a solid baseline that outperforms a raw model. To further enhance predictions, an auxiliary model similar to the first is trained, but excluding the timestep feature and using only non-healthy sequences with at least 10 rows. This model is designed to identify "Very High" risk rows (last two before failure), proving more accurate than direct "Medium" or "High" classification. The baseline classification for each sequence is then adjusted based on the maximum prediction score per sequence, using it as an indication of how close the last row is to failure.

Finally, a crucial insight that significantly enhanced the solution's effectiveness was the observation that features values very close to their minimum possible levels strongly indicated specific classifications. Additionally, it was observed that the training and test data, particularly the Gen2 test data, exhibited different distributions. Moreover, the test data contained a few  outliers with values much lower than the main distribution. To address these issues, the solution employed a normalization strategy that involved subtracting the 0.005 quantiles from each feature, applied separately to the training, Test-Gen1, and Test-Gen2 datasets. This approach effectively mitigated the impact of these discrepancies and outliers, leading to unexpectedly higher scores for Gen2 compared to Gen1.
\subsection{Team CarloMetta}

The proposed solution leveraged advanced ad-hoc machine learning techniques, specifically focusing on the use of Long Short-Term Memory (LSTM)\cite{lstm} networks combined with pseudo-labeling, boosting, and ensemble of different models.

The pipeline starts with a preprocessing phase where the training data is prepared to mirror the structure of the test set. This was critical to minimize discrepancies between training and testing environments and to simulate real-world conditions as closely as possible. Specific transformations included aligning the sequence lengths and ensuring the distribution of labels matched that of the test data scenarios.

Following data preprocessing, the core of our solution was built around a base LSTM model. This model was initially trained to predict risk levels at each timestep using the labeled training data.
The use of pseudo-labeling was pivotal in refining our model. After the base model generated predictions for the test set, we selected a subset of these predictions to serve as 'pseudo labels.' These were then added to the original training dataset, enhancing the training pool with new, diverse examples (from "gen2" trucks) that reflected test set characteristics. This technique not only helped refine the model's predictive accuracy but also helped adapt the model to generalize unseen data better.
To further optimize our model, we implemented a boosting technique through iterative training. Each iteration involved adding more layers to the LSTM network. The model started with simpler, shallower networks and progressively increased in complexity, culminating in a 10-layer deep LSTM network. Each network was trained on a dataset augmented by pseudo labels from previous iterations.

Our ensemble approach also played a crucial role in the final model deployment. By combining multiple models from different training iterations, we reduce individual model biases and enhance overall prediction stability. This ensemble not only served to validate the predictions through a majority voting mechanism but also ensured that our final output was more robust.

In our solution, we incorporated several ad-hoc techniques alongside our core LSTM and pseudo-labeling strategy to enhance model performance and robustness. One such technique was inspired by \cite{bias}, which explores the efficiency of adjusting neural network biases in place of the weights to improve generalization in deep learning models. We experimented with similar bias modifications in our LSTM layers to see if these adjustments could offer a more computationally efficient way to enhance prediction accuracy without substantially increasing the model's complexity. Additionally, we integrated the SwitchPath methodology, which involves dynamically altering the paths through which data flows in the network, as suggested in \cite{switchpath}. Although primarily developed for networks utilizing ReLU activations, we adapted its principles to suit the LSTM's typical tanh activation functions. This adaptation involved selectively activating different pathways during the training phase to avoid local minima and enhance the exploration of the model’s parameter space, potentially leading to more robust learning outcomes.

For post-processing, we focused on ensuring the logical consistency of the risk predictions across each sequence. Given the ordered nature of risk levels in predictive maintenance—where risk should not decrease as failure approaches—we adjusted sequences that did not adhere to this logic.

The integration of all these techniques formed the backbone of our solution. This approach meticulously addressed the challenge's demands, focusing on accurately predicting maintenance needs based on extensive real-world data. The code for this solution can be found at \url{https://github.com/CuriosAI/Volvo\_Discovery\_Challenge\_ECML\_PKDD\_2024}.

\subsection{Team MALTO (MAchine Learning at PoliTO)}
\label{sec:malto_solution}

\begin{figure}[h!]
    \centering
    \includegraphics[width=\linewidth]{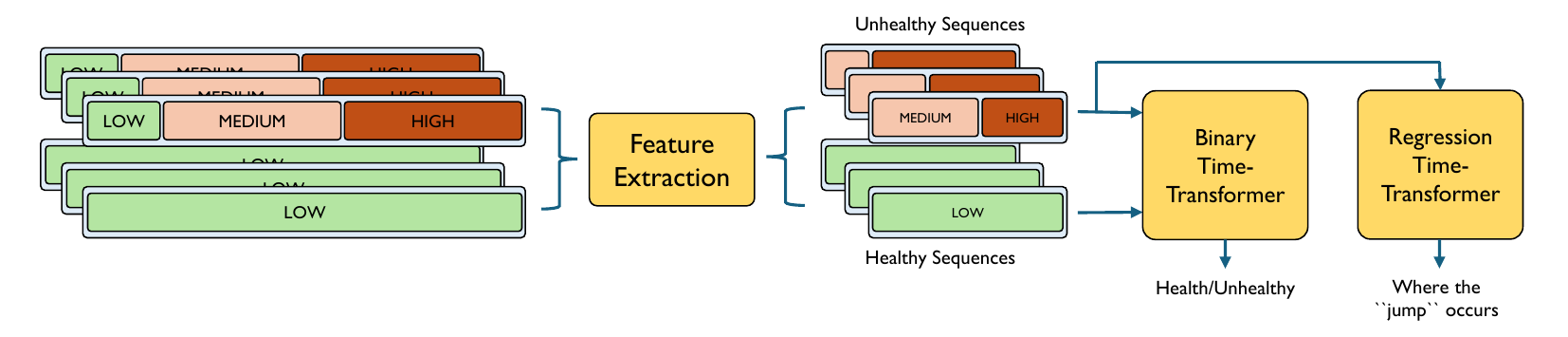}
    \caption{The training loop of the solution proposed by team MALTO: First, the feature extraction is done. Then, all the time series are resized to 10 timesteps following the criteria of the test set. Then, the training loop is done separately for the two models.}
    \label{fig:malto_solution}
\end{figure}
        
The proposed solution exploits the Time-Transformer \cite{liu2024time} in combination with different feature extraction techniques. The attention mechanism allows access to all historical input steps, focusing on the most relevant ones and extending the receptive field without increasing the depth of the network, one of the limitations of the normal TCN \cite{he2019temporal}. Fig. \ref{fig:malto_solution} shows the solution's pipeline overview. The solution's code can be found at \url{https://github.com/MAL-TO/Volvo-Discovery-Challenge-ECML-PKDD-2024}. 

To resume the data structure present in the test set, we sampled sequences of 10 timesteps from the training set. We sampled from the entire time series for the healthy data, while for the unhealthy data, we sampled starting from the first time step classified as "High" risk, going backward. To enhance the predictive performance of our model, we applied a series of feature extraction techniques to the temporal data. We computed the first derivative for the numerical features, which helps reveal trends and abrupt shifts in the data.
Additionally, we applied wavelet transforms to the numerical features, using a discrete wavelet transform (DWT) with the "db4" wavelet. This decomposition captures patterns across different time scales, offering insights into both high-frequency fluctuations and low-frequency trends. The resulting feature set includes the original numerical and categorical features, the first derivatives, and the wavelet-transformed data.
 
The proposed solution applies a two-step approach to correctly identifying the different risk levels "Low", "Medium", and "High". Two different Time-Transformers are used to solve the problem. The first one determines the healthy and unhealthy sequences as a binary classification problem. The second one considers only the non-healthy sequences and evaluates where the risk is "Medium" and "High". For this step, we introduce the definition of "jump", i.e. when the risk changes from “Medium” to “High”. We noticed that if we approach the problem as a sequence modeling task (i.e., classifying each timestep), the model had a hard time capturing some of the implicit rules of the solutions (e.g., always passing from "Medium" to "High" and not vice-versa, having only one jump per solution, etc.). To remedy this, we turned the problem into a regression task that would estimate when the jump would occur: by using a sigmoid activation on the output, we ensured that it would be contained in the range (0,1), and we interpreted it as the percentile of the time series where the jump took place (i.e., if the output is 0.65, then for a time series of length 10 the jump would take place between the 6th and the 7th timestep). In both models, the Time-Transformer is used as a feature extractor.  

\section{Terms and Conditions}
All participants of the Volvo Challenge ECML PKDD 2024 gained access to Volvo’s dataset. The dataset distributed during the challenge is referred to as the \texttt{Volvo GTT} dataset and belongs to Volvo. By participating in the challenge, Volvo granted the contestants and they accepted to receive a personal, non-exclusive, non-transferable, non-sublicensable, royalty free license to use the \texttt{Volvo GTT} dataset solely for the purpose of participating in the challenge. The license lasted only during the time of the Volvo Challenge ECML PKDD 2024. Apart from the abovementioned license, Volvo reserves all rights in the \texttt{Volvo GTT} dataset.

By enrolling in the competition, all participants granted Halmstad University (as the organizer) and Volvo Group a license to the solutions proposed in the contest. Given the prize levels, the contestants may be subject to income tax if they win, and participants were asked to carry any tax effects of receiving the prize.

\section*{Acknowledgements}

We would like to acknowledge the Swedish AI Society for their valuable contributions and support.


\begin{thebibliography}{8}








\bibitem{voskou2024transformers}
Andreas Voskou, Charalambos Christoforou, and Sotirios Chatzis.
\newblock Transformers with stochastic competition for tabular data modelling.
\newblock In {\em ICML 2024 Workshop on Structured Probabilistic Inference \& Generative Modeling}.

\bibitem{he2019temporal}
He, Y., Zhao, J.: Temporal convolutional networks for anomaly detection in time series. In: Journal of Physics: Conference Series, vol. 1213, no. 4, pp. 042050. IOP Publishing (2019)

\bibitem{liu2024time}
Liu, Y., Wijewickrema, S., Li, A., Bester, C., O'Leary, S., Bailey, J.: Time-Transformer: Integrating local and global features for better time series generation. In: 2024 SIAM International Conference on Data Mining (SDM), pp. 325--333. SIAM (2024)


\bibitem{rahat2023bridging}Rahat, M., Kharazian, Z., Mashhadi, P., Rögnvaldsson, T. \& Choudhury, S. Bridging the Gap: A Comparative Analysis of Regressive Remaining Useful Life Prediction and Survival Analysis Methods for Predictive Maintenance. {\em PHM Society Asia-Pacific Conference}. \textbf{4} (2023)

\bibitem{lstm}
Hochreiter, S. et alt.: Long Short-Term Memory. In: 1997 Neural Computation, pp. 1735-1780, vol 9, number 8.

\bibitem{bias}
Metta, C. et alt.: Increasing biases can be more efficient than increasing weights. In: 2024 IEEE/CVF WACV.

\bibitem{rahat2024survloss}Rahat, M. \& Kharazian, Z. SurvLoss: A New Survival Loss Function for Neural Networks to Process Censored Data. {\em PHM Society European Conference}. \textbf{8}, 7-7 (2024)

\bibitem{switchpath}
Di Cecco, A. et alt.: SwitchPath: Enhancing Exploration in Neural Networks Learning Dynamics. In: 2024 Discovery Science.

\bibitem{altarabichi2020stacking}Altarabichi, M., Mashhadi, P., Fan, Y., Pashami, S., Nowaczyk, S., Moral, P. \& Rognvaldsson, T. Stacking ensembles of heterogenous classifiers for fault detection in evolving environments. {\em 30th European Safety And Reliability Conference, Esrel 2020 And 15th Probabilistic Safety Assessment And Management Conference, Psam15}. pp. 1068-1068 (2020)


\end{thebibliography}
\end{document}